\pdfoutput=1
 
\documentclass[11pt]{article}
\usepackage[dvipsnames]{xcolor}

\usepackage{url}
\usepackage[breaklinks]{hyperref}
\usepackage{emnlp2022}
\usepackage{enumitem}

\usepackage{times}
\usepackage{latexsym}
\usepackage{booktabs}
\usepackage{adjustbox} 
\usepackage{tabularx}

\usepackage{multirow}
\usepackage{physics}
\usepackage{cleveref}
\usepackage{caption}
\usepackage{arydshln}
\usepackage{subcaption}
\crefname{equation}{Eq.}{Eq.}
\crefname{section}{Sectio
n}{Sections}
\crefname{subsection}{Section}{Sections}
\crefname{subsubsection}{Section}{Sections}
\crefname{figure}{Figure}{Figures}
\crefname{table}{Table}{Tables}
\crefname{subfigure}{Figure}{Figures}
\crefname{algocf}{Algorithm}{Algorithms}

\usepackage[T1]{fontenc}

\usepackage[utf8]{inputenc}
\newcommand{\red}[1]{\textcolor{BrickRed}{#1}}

\newcommand{\green}[1]{\textcolor{Green}{#1}}
\newcommand{\yellow}[1]{\textcolor{Goldenrod}{#1}}
\usepackage{microtype}
\usepackage{multirow}
\usepackage[makeroom]{cancel}
\usepackage[autostyle]{csquotes} 
\usepackage[english]{babel}
\usepackage{graphicx}
\usepackage{amsmath}
\DeclareMathOperator*{\argmax}{arg\,max}

\newcommand{\newpara}[1]{\vspace{0.3em} \noindent \textbf{#1}~ \hspace{0.4em}}
\title{Adjust the Belief: an Interpretable Debias with Constraint Optimization 
}
\title{
Fairness $\neq$ Blindness: 
Using
Sensitive Information \\with Interpretable Fairness Constraints}

\title{
Fairness $\neq$ Blindness: 
Debias with Interpretable Fairness Constraints under Sensitive Information Exposed}

\title{
Debiasing While Allowing Sensitive Information\\ Within the Boundaries of Interpretable Fairness Contraints}

\title{Controlling Bias Exposure for Fair Interpretable Predictions}

\author{Zexue He, Yu Wang }
\author{Zexue He, Yu Wang, Julian McAuley, Bodhisattwa Prasad Majumder \\
  Department of Computer Science and Engineering\\
  University of California, San Diego  \\
  \texttt{\{zehe@eng, yuw164@, jmcauley@eng, bmajumde@eng\}.ucsd.edu}}

\begin{document}
\maketitle
\begin{abstract}
Recent work on reducing bias in NLP models usually focuses on protecting or isolating  information related to a sensitive attribute (like gender or race). However, when sensitive information is semantically entangled with the task information of the input, e.g., gender information is predictive for a profession, a fair trade-off between task performance and bias mitigation is difficult to achieve. 
Existing approaches perform this trade-off by eliminating bias information from the latent  space, lacking control over how much bias is necessarily required to be removed.
We argue that
a favorable debiasing method should use sensitive information `fairly', rather than blindly eliminating it 
\cite{caliskan2017semantics,sun2019mitigating, bogen2020awareness}
.
In this work, we provide a novel debiasing algorithm by adjusting the predictive model's belief to (1) ignore the sensitive information if it is not useful for the task; (2) use sensitive information \emph{minimally} as necessary for the prediction (while also incurring a penalty).
Experimental results on two text classification tasks (influenced by gender) and an open-ended generation task (influenced by race)  
indicate that our model achieves a desirable trade-off between debiasing and task performance along with producing debiased rationales as evidence.


\end{abstract}

\section{Introduction}
Human-written language contains implicit or explicit biases and stereotypes, which make their way into deep natural language processing (NLP) systems through the learning procedure. Emerging works show that biases may have worrisome influence and even lead to unfair outcomes in various NLP tasks like text classification \cite{park2018reducing, kiritchenko2018examining,de2019bias}, coreference resolution \cite{rudinger2018gender}, toxicity detection \cite{zhou2021challenges,xia2020demoting, xu2022leashing}, language modeling \cite{lu2020gender, bordia2019identifying, sheng2019woman}, etc.

Recently, several works have attempted to address bias issues in NLP tasks. One stream of approaches is sensitive attribute protection \cite{zhang2018mitigating, jentzsch2019semantics, badjatiya2019stereotypical, heindorf2019debiasing, he2021detect}, which mitigates bias by isolating or protecting certain sensitive attributes like race or gender from decision making. However, real-world human-written language is complicated and there are often cases where sensitive information is entangled tightly with the semantics of the sentence \cite{caliskan2017semantics}. 
In this situation, protecting the attribute will unavoidably affect the model’s performance. For example, isolating all the underlined words in
\begin{displayquote}
Example 1.
\textit{ \underline{He} is a \underline{congressman} and \underline{he} is good at singing. }
\label{exp: intro}
\end{displayquote}
might misguide a `profession' classifier to get a result of a  \textit{singer} (instead of a \textit{congressman}). 
The balance between bias mitigation and other desired goals is challenging in current debiasing scenarios \cite{sheng2021societal}. 
Conceptually, debias methods that protect sensitive attributes in some latent space may achieve such a delicate equilibrium if bias is reduced to some precise degree. However, controlling the degree of debiasing in a transparent fashion is challenging \cite{gonen2019lipstick} as these methods \cite{zhang2018mitigating,ravfogel2020null, gonen2019lipstick} operate in a black-box style, providing no evidence for bias mitigation or task performance. Hence, it remains hard for human users to understand and trust the underlying debiasing mechanism.
\begin{figure*}
  \includegraphics[width=\textwidth]{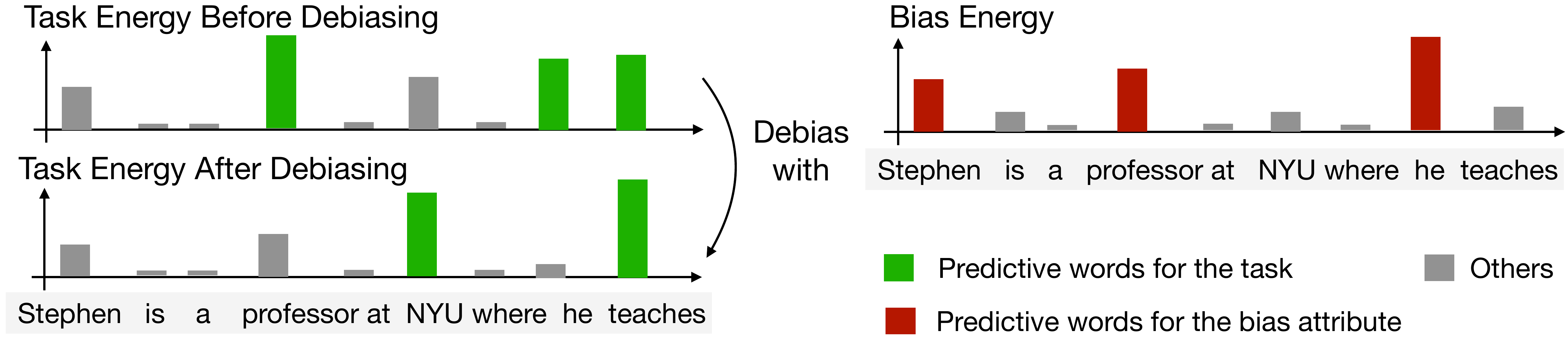}
  \caption{ \textbf{Example} of how our debiasing algorithm works. We regulate the contribution (energy) of each token responsible for `profession' classification according to their predictability of `gender'. Task energy of a biased token is decreased and re-allocated to its replacement.
  }
  \label{fig: constraint}
\end{figure*}

Inspired by 
\citet{caliskan2017semantics}, we believe a favorable debiasing method should aim to teach a model to behave fairly instead of blinding its perspective from certain sensitive information \cite{sun2019mitigating, bogen2020awareness}.  To this end, we propose a novel debiasing algorithm  %
that produces evidence behind a task prediction while constraining the evidences as much bias-free as possible. 

We design our algorithm based on following principles: it is fair to (1) ignore a sensitive information if it is not useful for the task prediction; (2) use a minimal amount of sensitive information if they are necessary for the task. In \cref{fig: constraint}, we can find that our method identifies `professor' is often predictive of gender and is not necessary to be used for predicting profession when there are other useful non-biased words such as `NYU', `teaches' etc. We aim to achieve two goals: a desired and fair balance between task performance and bias mitigation, and producing debiased rationales as an evidence for the task prediction.


Recent works \cite{lei-etal-2016-rationalizing, bastings-etal-2019-interpretable} have shown that \emph{rationales} are an effective way to justify the reasoning behind a prediction from a neural model. Therefore, we work with rationales for task prediction and measure their importance based on \emph{energy} for both task prediction and being biased. We eventually optimize the task rationale in such a way that all tokens of the task rationales will have low bias energy without sacrificing the task performance by blindly removing all bias information.

We evaluate our method on two classification tasks that are influenced by gender and an oped-ended generation task that is influenced by race as a sensitive attribute.
Comprehensive experiments reveal that our method achieves best trade-off between task performance and bias mitigation, simultaneously producing concise and faithful rationales. We indeed observe that extreme debiasing in baselines hurt task performance whereas performance-aware removal of sensitive information does not affect model performance, rather improves interpretability. To the best of our knowledge, our work is the first to investigate debiasing using interpretable models and we hope that this work will provide a new perspective of controllable debiasing for fair interpretable models. Our codes are released in \url{ https://github.com/ZexueHe/interpretable_debiasing}.

\section{Related Work}
\paragraph{Debiasing on Data} is a debiasing method that focuses on augmenting or cleaning the existing datasets. Counterfactual Data Augmentation (CDA) \citet{lu2020gender} replaces the bias component of each example in a dataset with a counterfactual one. Several works followed CDA to propose specific augmentation functions for Coreference Resolution \cite{zhao-etal-2018-gender}, Machine Translation \cite{saunders-byrne-2020-reducing,costa2020fine}, Language Modeling \cite{sheng2019woman}. Despite being effective, CDA's augmenting functions are heuristic and require human intervention. Data cleaning for debiasing aims to generate a neutral version of biased input with paraphrasing techniques such as back-translation \cite{xu2019privacy} and rewriting \cite{he2021detect}, however it is often challenging to maintain the same semantic meaning before and after paraphrasing.

\paragraph{Debiasing on Representation} methods usually operate on the embedding space of inputs \cite{lu2020gender,dathathri2019plug} or tokens \cite{escude-font-costa-jussa-2019-equalizing, caliskan2017semantics, zhao-etal-2018-gender,bolukbasi2016man}. The sensitive information is removed by optimizing the encoder with reversed gradients from a bias discriminator \cite{zhang2018mitigating,dathathri2019plug}, or projecting the latent space to an orthogonal subspace \cite{ravfogel2020null,subramanian2021evaluating}. Some works also design the regularization techniques for equalizing bias-specific tokens \cite{zhao2018learning, bolukbasi2016man}. However, these methods are typically black-box, and controlling the degree of debiasing is often difficult without affecting the task performance \cite{gonen2019lipstick}.  

Our method aims to understand bias in predictive models and mitigate it while maintaining task performance in a controllable and interpretable fashion. In general, our method does not contradict previous works in terms of debiasing, and can be flexibly combined with other debiasing methods (e.g.,~ CDA first, then ours).

\section{Approach}
In this section, we introduce our interpretable debiasing algorithm that uses a `fair' amount of sensitive information in the important parts of input (a.k.a. rationale).
We aim to perform a predictive task (e.g.,~predicting a profession based on a biography) while minimizing the impact of sensitive information (e.g.,~gender) with minimally affecting the  performance of the original task. 
Given an input, there are tokens that are predictive of the task output (we call them task rationales) and there are tokens that carry the sensitive information (we call them bias rationales). With energy functions, we measure how important a token is for the task output or how sensitive it is. By constraining the use of biased input tokens, we control the task energy so that the model is allowed to be exposed to a minimum of bias that is necessary to the task.

%

%

\begin{figure}
  \includegraphics[width=\linewidth]{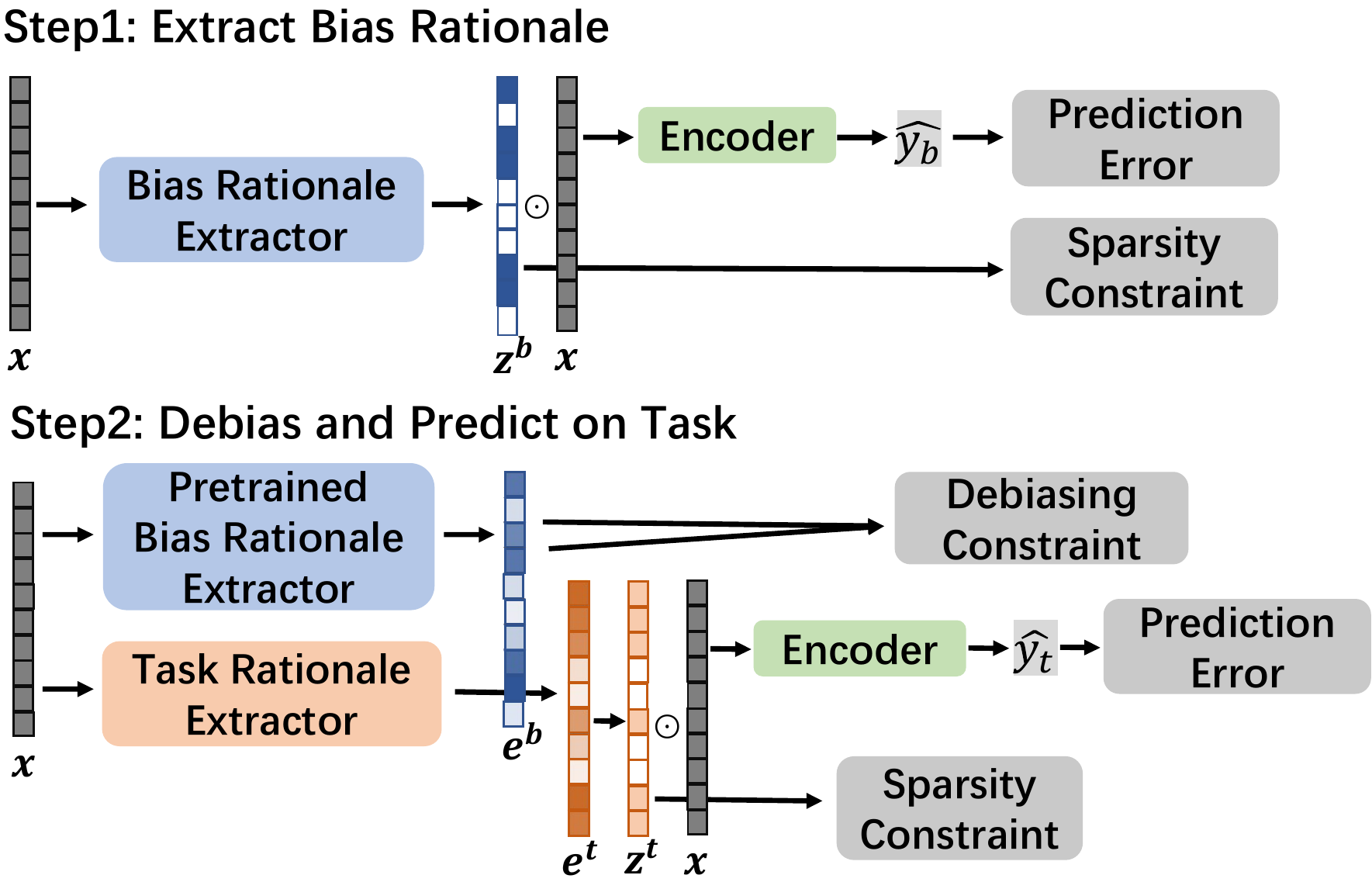}
  \caption{ \textbf{Pipeline}. We first pretrain a bias rationale extraction framework and obtain bias energy for each input token. Then we train a \emph{fair} task prediction model where the task rationales are regulated by a debiasing constraint based on bias energy. A token with high bias energy will be penalized for being in task rationale with a decrease in its original task importance.
  }
  \vspace{-0.8em}
  \label{fig: model}
\end{figure}

\subsection{Extracting Bias Rationale}
We first identify input tokens that carry sensitive information. To be more specific, for an input text $\mathbf{x} = \{x_1, x_2, x_3, \cdots, x_n\}$ with $n$ tokens (e.g.,~biography of a person), 
we predict the bias label $y_{b}$ (e.g.~gender of the person, having $K_b$ categories) based on $\mathbf{x}$ with model $f_b(\mathbf{x};\theta_b)$ parameterized by $\theta_b$, so that the predicted bias label $\hat{y_{b}}$ is close to ground truth $y_{b}$
\begin{equation*}
    \hat{y_{b}} = \argmax_{k_b\in K_b} f_b(\hat{y_b}=k_b|\mathbf{x}; \theta_b),
\end{equation*}
which is optimized by minimizing the cross-entropy error $\mathcal{L}_{bias}(f(\mathbf{x}), y_{b};\theta_b)$. 
We are interested in identifying the tokens that are most predictive for $\hat{y_{b}}$, i.e. bias rationales.

Rationale is defined as a short yet sufficient snippet of an input responsible for the prediction \cite{bastings-etal-2019-interpretable}. 
Here, we obtain the bias rationale using an extractive framework that includes two modules -- an extractor that identifies parts of input as the rationale, and an encoder that makes a prediction only based on the rationale. The extractor and encoder together compose the rationale extraction framework (REF). The proposed rationale comes in the form of a sequence of binary variables, indicating if a particular input token is informative to the task. The extractor and the encoder are jointly trained to minimize the prediction error.

Therefore, to extract bias rationale, 
we augment $f_b$ with the sequence of latent binary variables $\mathbf{z^b}=\{z^b_1, z^b_2, z^b_3, \cdots, z^b_n\}$, $z^b_i \in \{0, 1\}$ \cite{lei-etal-2016-rationalizing}, 
which is optimized to maximize the predictive probability of the correct bias label by regulating the contribution of each token:  
\begin{align*}
    \mathbf{z^b} &\sim g_b(\mathbf{x}|\phi_b) \\
    \hat{y_b} &=\argmax_{k_b\in K_b} f_b(y_b=k_t | \mathbf{x}\odot \mathbf{z^b} ; \theta_b)
\end{align*}
where $g_b$ is a bias rationale extractor parameterized by $\phi_b$, that predicts the probability of how much each token contributes to predict the bias label. We sample the binary vector $\mathbf{z^b}$ from $g_b$ and  $\mathbf{x} \odot \mathbf{z^b}$ is treated as the \emph{bias rationale}. We model $g_b$ such that the output of $g_{b}$ satisfies Kuma distribution \cite{bastings-etal-2019-interpretable} 
to avoid $\mathbf{z^b}$ being non-differentiable.
Bias REF is trained with the following objective and important tokens for predicting bias are selected as bias rationales:
\begin{align*}
    \mathcal{C}_{b} = \mathcal{L}_{b}(f_b(\mathbf{x}\odot\mathbf{z^b}); \theta_b) + \lambda_b \Omega_{b}(\phi_b)
\end{align*}
where $\lambda_b$ is hyperparameter and $\Omega_b$ is a sparsity constraint penalizing the number of selections and translations, making learned rationale concise and sufficient.


\subsection{Task Prediction}
Based on the bias rationale obtained so far, we want to influence a predictive model to use input tokens in a debiased way. Elaborately, we want the contribution of the biased tokens to be as minimal as possible for the predictive task. To achieve this, we encourage the predictive model for a task (e.g.,~profession classification with $K_t$ classes) to use informative tokens (task rationales) with minimal bias.


Similar to bias rationale extraction, we train a task REF consists of an extractor $g_t$ that generates $\mathbf{z_t}= [z_1^t, z_2^t, \cdots, z_3^t]$, and an encoder $f_t$ that makes prediction with extracted rationale $\vb{x} \odot \vb{z}^t$\:
\begin{align}
    \vb{z^t} &\sim g_t(\vb{x}| \phi_t) \nonumber \\
    \hat{y}_t & = \argmax _ {k \in K_{t}} f_t(\hat{y}_t=k_t| \vb{x} \odot \vb{z}^t ; \theta_t)\nonumber
\end{align}
where $\hat{y}_t$ is the task prediction and $y_t$ is the ground truth label ($y_t \in C_t$). Task rationale is extracted by minimizing the task cross-entropy loss $\mathcal{L}_{t}$ and maintaining the sparsity $\Omega_{t}$, as 
\begin{align*}
    \mathcal{C}_{t} = \mathcal{L}_{t}&(\mathcal{F}(\mathbf{x}\odot\mathbf{z^t}); \theta_t) + \lambda_t \Omega_{task}(\phi_t)
\end{align*}

However, we would like to modify the task REF to consider bias rationale, and optimize task rationale in such a way that they contain minimal bias. For this, we introduce a debiasing constraint that adds a penalty if a biased token is used as the part of the task rationale, and optimize the task rationale to incur minimal penalty.

\subsection{Debiasing with Energy-Based Constraint}
\label{sec: constraint}


Our debiasing constraint should regulate the importance of the biased tokens towards the predictive task.
We capture the importance of each token for being biased and being important for the predictive task, using \emph{energy} scores\footnote{We did not use direct probabilities from REFs since they produce unstable performance as $p(z^b_i=0)$ and $p(z^t_i=0)$ may not be independent and may not be summable. See~\cref{sec: experiment} for the experimental evidences. 
}.
\emph{Energy} is defined as the negative log-likelihood of the non-selection probability of each token \citep{lecun2006tutorial}. Higher energy indicates stronger importance.

We obtain the task energy for the $i$-th token as:
\begin{align*}
    e_i^{t} &= -\operatorname{log-likelihood}(p(z^t_i = 0)) \nonumber \\
    &= -\operatorname{log-likelihood}(1 - g_t(x_i|\phi_t)),
\end{align*}
where $g_t(x_i|\phi_t)$ is the probability for selecting the $i$-th token $x_i$ for the task prediction. Similarly, the bias energy for the $i$-th token would be:
\begin{align*}
    e_i^{b} &= -\operatorname{log-likelihood}(1 - g_b(x_i|\phi_b))
\end{align*}



We construct the debiasing constraint using both task and bias energy for a token. For an $i$-th token that has a high bias energy, we will penalize its importance for the predictive task by decreasing its task energy. In contrast, for tokens with low bias energy, we keep  their task energy as it is. This is realized by a debiasing constraint as:
\begin{align}
    D(i)= \left \{
    \begin{aligned}
     e_i^t+ (e_i^b&-A) &\text{ if } e_i^b > A, \nonumber\\
     &0 &\text{otherwise}
    \end{aligned}
    \right .
    \label{eq: constraint}
\end{align}
where $A$ is a hyperparameter indicating the bias tolerance threshold \footnote{Setting the threshold to the minimum of bias energy values will result in removing all biased tokens, prohibiting using any sensitive information.}.
This constraint will eventually get rid of highly biased token for being important to the task and use low-bias energy replacements instead, in order to boost the task performance. This modifies our task objective as:
\begin{equation*}
    \mathcal{C} = \mathcal{C}_{t} + \gamma\sum_i^{|\vb{x}|} D(i)
\end{equation*}
where $\gamma$ is the hyperparameter.  

\subsection{Training}
The pipeline of our algorithm is shown in \cref{fig: model}.
We first pretrain a bias REF $f_b$ by minimizing $\mathcal{C}_b$. During the debiasing process, this model is served as a fixed reference model. During debiasing, we then train the task model $f_t$ by minimizing $\mathcal{C}$. 
For classification tasks, $\mathcal{L}_{t}$ is a cross-entropy loss and for generation task, $\mathcal{L}_{t}$ is a language-modeling loss. Hyperparameters and more details on training are provided in Appendix~\ref{sec: Hyperparameter Study}.



\section{Experimental Setup}
\label{sec: experiment}






\begin{table*}[t!]
\small
\centering
\resizebox{0.9\linewidth}{!}{%
\begin{tabular}{ccccccc}
\toprule 
 Task & 
  \bf Variants &
  \begin{tabular}[c]{@{}c@{}}\textbf{Toxicity} \\ \textbf{F1 Score}  \end{tabular} $\uparrow$ &
  \begin{tabular}[c]{@{}c@{}}\textbf{Gender}\\ \textbf{F1 Score}  \end{tabular} $\downarrow$ &
  \begin{tabular}[c]{@{}c@{}}\textbf{Comprehensive-}\\ \textbf{ness Score}\end{tabular} $\uparrow$ &
  \begin{tabular}[c]{@{}c@{}}\textbf{Sufficiency}\\ \textbf{Score}  \end{tabular}$\downarrow$ &
  \textbf{Selection}$\downarrow$ \\ \midrule
\multirow{4}{*}{\begin{tabular}[c]{@{}c@{}}\textbf{Toxicity} \\ \textbf{Detection}\end{tabular}}        & Full Text   & 0.73 & 0.56 & -       & -       & 100\%   \\
                                                                                      & Reranking   &   0.64     &  0.39      &  0.01       &      0.01   &  34.7\%      \\
                                                                                      & Probability & 0.65 & 0.37 & 0.00 & 0.00 & \bf 63.42\% \\
                                                                                      & Ours        & 0.73 & \bf 0.37 & \bf 0.00  & \bf 0.00 & \bf 63.34\% \\ \cmidrule{0-6}
 Task &
  \bf Variants &
  \begin{tabular}[c]{@{}c@{}}\textbf{Profession} \\ \textbf{Accuracy} \end{tabular} $\uparrow$ &
  \begin{tabular}[c]{@{}c@{}}\textbf{Gender}\\ \textbf{F1 Score} \end{tabular} $\downarrow$ &
  \begin{tabular}[c]{@{}c@{}}\textbf{Comprehensive-}\\ \textbf{ness Score} \end{tabular}$\uparrow$ &
  \begin{tabular}[c]{@{}c@{}}\textbf{Sufficiency}\\ \textbf{Score}\end{tabular}$\downarrow$ &
  \textbf{Selection} $\downarrow$ \\ \cmidrule{0-6}
\multirow{4}{*}{\begin{tabular}[c]{@{}c@{}}\textbf{Profession} \\ \textbf{Classification}\end{tabular}} & Full Text   & 0.81 & 0.98 & -       & -       & 100\%   \\
                                                                                      & Reranking   & 0.70        &      0.45  &      0.23   &        0.32 & 36.40\%       \\
                                                                                      & Probability & 0.73 & 0.50 & 0.44  & 0.13  & \bf 65.42\% \\
                                                                                      & Ours        & 0.80 & \bf 0.38 & \bf 0.52  & \bf 0.01  & \bf 65.26\% \\ \bottomrule
\end{tabular}
}
\caption{ Evaluation of rationale-based debiasing methods on classification tasks}
\label{tab:result of classification variant }
\end{table*}

\begin{table}[t!]
\small
\centering
\resizebox{0.7\linewidth}{!}{%
\begin{tabular}{ccc}
\toprule
\bf Models     & \bf Toxicity F1 $\uparrow$ & \bf Gender F1 $\downarrow$\\ \midrule
Full Text & 0.73 & 0.56 \\ \midrule
Adv   & 0.46      & 0.22    \\
Embed & 0.49      & 0.30    \\
Ours   & 0.73      & 0.37    \\ \bottomrule
\end{tabular}%
}
\caption{ Comparison between ours  and other debiasing baselines without rationales on toxicity detection}
\label{tab:jigsaw debiasing baselines}

\end{table}

\begin{table}[t!]
\centering
\resizebox{\linewidth}{!}{%
\begin{tabular}{cccc}
\toprule
\bf Models    & \bf Profession Acc. $\uparrow$ & \bf Gender F1 $\downarrow$ & \bf RMS TPR-GAP$\downarrow$\\ \midrule
Full Text & 0.813           & 0.984     & 0.184       \\ \midrule
Adv      & 0.361           & 0.358     & 0.057       \\
INLP  & 0.752           & -         & 0.095       \\
Embed    & 0.236           & 0.914     & 0.179       \\
Ours      & 0.796           & 0.375     & 0.054       \\ \bottomrule
\end{tabular}%
}
\caption{Comparison between ours and other debiasing baselines without rationales on profession classification}
\label{tab:biobias debiasing baselines}
\vspace{-0.2em}
\end{table}

\subsection{Scenarios and Datasets}
We evaluate our debiasing algorithm on two text classification tasks influenced by \emph{gender} bias --toxicity detection and profession classification, and an open-ended text generation task influenced by \emph{racial} bias. 
We use the Jigsaw Toxicity dataset \footnote{https://www.kaggle.com/c/jigsaw-unintended-bias-in-toxicity-classification} for toxicity detection, BioBias dataset \cite{de2019bias} for profession classification, and BOLD dataset \cite{dhamala2021bold} for open-ended generation.

\newpara{Jigsaw Toxicity} is a dataset for the Kaggle Toxic Comment Classification Challenge that detects toxicity (toxic or non-toxic) from a conversational response influenced by  multiple sensitive attributes. A datapoint has an input as a textual comment associated with annotated toxicity labels and various identity attributes about the entity mentioned, such as gender, race, etc. We take gender identification as the unintended bias and filter out the examples annotated as `no gender mentioned.' The gender categories in our dataset are female, male, transgender, and other gender. We have 125,071 examples out of which 80\%, 10\% and 10\% are used for training, validation, and testing respectively. 

\newpara{BiosBias} is a dataset derived from a large-scale user study of gender in occupation classification \cite{de2019bias}. It consists of short biographies annotated with gender and occupation information. \citet{de2019bias} found possible influence of gender behind the annotated profession labels. We consider a profession classification task without the influence of gender. We follow the experimental settings in \citep{ravfogel2020null}, that contains 393,423 biographies labeled with binary gender (male/female) and 28 professions (e.g.~professor, software engineer, model, etc.). 255,710 examples (65\%) are used for training, 39,369 (10\%) for validation, and 98,344 (25\%) for testing.

\newpara{BOLD} or Bias in Open-ended Language Generation Dataset is proposed by \citet{dhamala2021bold} to measure the fairness in open-ended language generation. This dataset contains 23,679 text generation prompts related to five domains: profession, gender, race, religious ideologies, and political ideologies, with corresponding ground-truth sentences taken from English Wikipedia. We divide the finetune/development/test set of examples in each domain with a 0.7/0.1/0.2 ratio, which is used to finetune a GPT2 language model. We then consider the four races (European Americans, African Americans, Asian Americans, and Latino/Hispanic Americans) as unintended bias. This subset consists of 7,657 prompts and ground truth, of which 5,359 (70\%) are finetuning examples, 765 (10\%) are validation examples, and 1530 (20\%) are test examples.


\newpara{Toxicity detection.} We first consider a baseline with full text input for toxicity detection. It provides the upper bound for task performance while still being mostly biased. We also consider two other debiasing methods as baselines: a model with adversarial training (Adv.) \cite{zhang2018mitigating} that performs debiasing on the model's latent space, and a model \citep{bolukbasi2016man} that performs debiasing on the embedding space (Embed). 

\newpara{Profession classification.}
Similar to toxicity detection, we also have the baseline with full text input that gives the upper bound of task performance but with maximum bias. For debiasing baselines we have Adv \cite{zhang2018mitigating} and INLP \cite{ravfogel2020null}, a method\footnote{Due to unavailability of the codes for INLP, gender prediction performance is not reported in \cref{tab:biobias debiasing baselines}. We use similar data settings as INLP to make other results comparable.} that removes bias with an iterative null-space projection. 


\begin{table}[t!]
\small
\centering
\begin{tabular}{c|cc}
\toprule
\bf Input                                                         & \bf Toxicity F1 & \bf Gender F1 \\ \midrule
Full Text                                                     & 0.73      & 0.56    \\ 
Toxicity Rationale  & 0.73      & 0.55    \\ \cmidrule{1-3}
Difference $\Delta$                                                      & 0.00      & 0.01    \\ \bottomrule
\end{tabular}
\caption{Toxicity and gender prediction with various inputs}
\label{tab: entangle jigsaw}
\end{table}

\begin{table}[t!]
\small
\centering
\begin{tabular}{c|cc}
\toprule
\bf Input                                                         & \bf Profession Acc. & \bf Gender F1 \\ \midrule
Full Text                                                     & 0.81              & 0.98    \\
Toxicity Rationale & 0.80              & 0.98    \\\cmidrule{1-3}
Difference $\Delta$                                                      & 0.01              & 0.00    \\ \bottomrule
\end{tabular}
\caption{ Profession and gender prediction with various inputs}
\label{tab: entangle biobias}
\vspace{-1.5em}
\end{table}

\newpara{Open-ended Generation.} We consider a language model (GPT2) trained on the original data to provide the upper bound of generation performance but with maximum bias. For debiasing baseline, we compare with PPLM  \cite{dathathri2019plug}, a controllable text generation algorithm which generates output by steering the generation away from the sensitive information. 



\newpara{Ablations.} To investigate the impact of different parts of our algorithm, we also considered two variants for comparison: (1) \emph{Rerank} where the task rationale is selected based on a reversed order of bias energy. This is an inference-time debiasing method, which is used to investigate the necessity of debiasing constraint during training (2) \emph{Probability} where we use probability directly obtained from REFs instead of energy for token importance. 

\newpara{Backbone Models.} In implementation, we use LSTM as the backbone for REFs in toxicity detection and profession classification, and use GPT-2 transformer as the backbond model in open-ended generation. See \cref{sec: appendix} for more details.

\subsection{Evaluation Metrics}
To ensure the optimal trade-off between bias removal and task performance we evaluate our model based on three desiderata: (1) task performance, (2) bias mitigation, and (3) rationale faithfulness.  

\newpara{Task Performance.} 
To evaluate task performance, we use F1 scores for toxicity prediction due to the imbalanced output label proportions and use accuracy for profession classification. For the open-ended generation task, the goal is to generate a high-quality sentence following a prompt. We use language model perplexity and BertScore \cite{zhang2019bertscore} w.r.t. the ground-truth text.

\begin{table*}[t!]
\centering
\resizebox{0.95\linewidth}{!}{%
\begin{tabular}{ccccccccc}
\toprule
 &
  \bf Models &
  \textbf{PPL}$\downarrow$ &
  \begin{tabular}[c]{@{}c@{}}\textbf{BertScore} \\ \textbf{Precision}  \end{tabular} $\uparrow$&
  \begin{tabular}[c]{@{}c@{}}\textbf{BertScore}\\ \textbf{Recall}  \end{tabular} $\uparrow$&
  \begin{tabular}[c]{@{}c@{}}\textbf{BertScore}\\ \textbf{F1}  \end{tabular} $\uparrow$&
  \begin{tabular}[c]{@{}c@{}}\textbf{Race}  \\ \textbf{Accuracy} \end{tabular} $\downarrow$&
  \begin{tabular}[c]{@{}c@{}}\textbf{Sufficiency}  \\ \textbf{Score} \end{tabular} $\downarrow$&
  \textbf{Selection} $\downarrow$ \\ \midrule
\multirow{6}{*}{\begin{tabular}[c]{@{}c@{}}\textbf{Open-ended} \\ \textbf{Generation}\end{tabular}} 
                                                                 & Ground Truth & 27.69 & 1.00 & 1.00 & 1.00 & 0.63 & - & 100.0\% \\
                                                                 & GPT2     & 69.61    & 0.86   & 0.86 & 0.86 &     0.62   &  41.92 & 60.2\%  \\
                                                                                & PPLM     & 66.97        & 0.81 & 0.81 & 0.81        & 0.61 & 39.28 &100.0\%          \\
               & Rerank     & 69.73        & 0.84 & 0.85 & 0.85        & 0.62& 42.04 & 37.7\%         \\
                                                                              
                                                                               & Probability  & 77.69       & 0.88 & 0.87 &    0.87    & 0.62 & 50.00 &53.7\% \\
                                                                               & Ours         &   67.22 &  0.86 & 0.86 & 0.86 &  0.62 & 39.51& 51.9\% \\ \bottomrule
\end{tabular}
}
\caption{ Comparision of our method with debiasing baselines on open-ended generation task} 
\vspace{-0.8em}
\label{tab: Bold}
\end{table*}
\begin{figure}
     \centering
     \begin{subfigure}{0.493\linewidth}
         \centering
         \includegraphics[width=\linewidth]{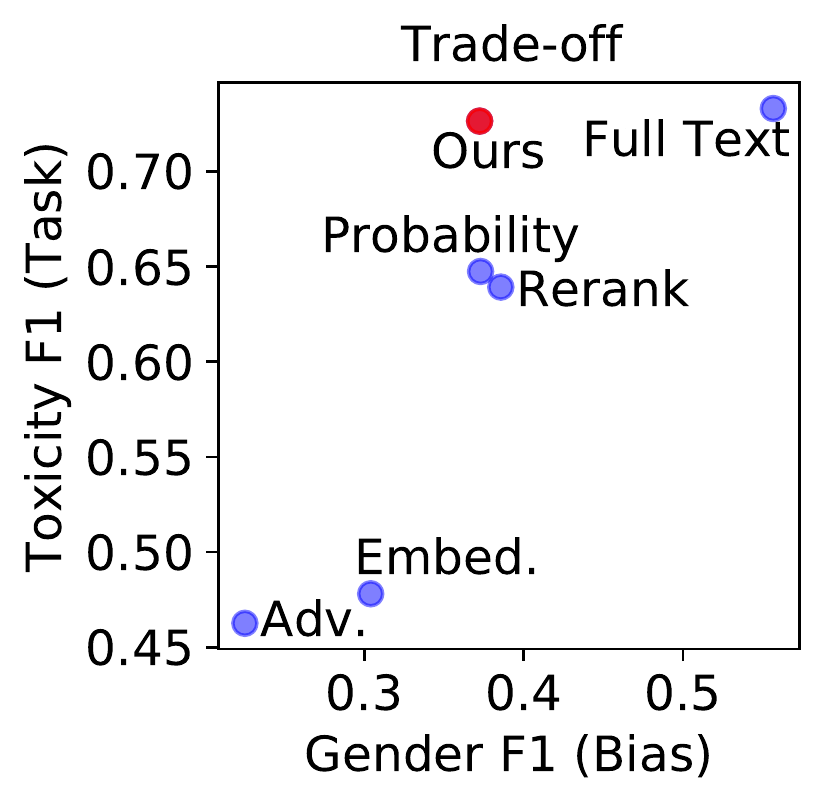}
         \vspace{-1.7em}
         \caption{}
         \label{fig: toxicity}
     \end{subfigure}
     \hfill
     \begin{subfigure}{0.493\linewidth}
         \centering
         \includegraphics[width=\linewidth]{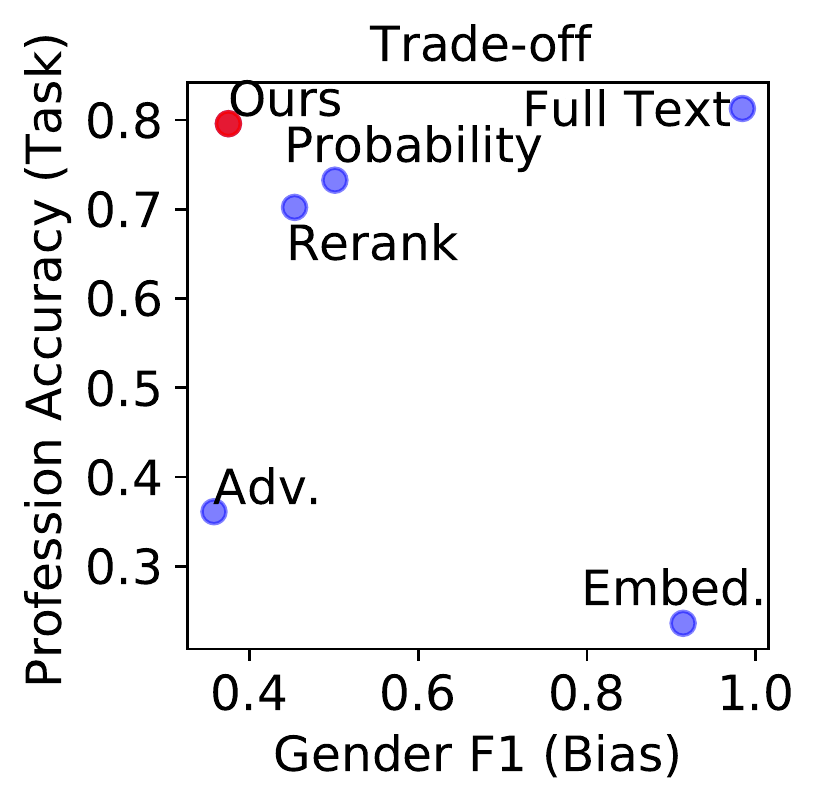}
         \vspace{-1.7em}
         \caption{}
         \label{fig: profession}
     \end{subfigure}
     \vspace{-0.5em}
     \caption{Trade-off between bias and task performance for (a) Toxicity Detection (b) Profession Classification.  More upper left means a better model.}
     \label{fig: trade off in text classifition}
     \vspace{-1em}
    \end{figure}

\subsection{Baselines and Ablations}
\label{sec:baselines}

\newpara{Bias Mitigation.} Following \citet{zhang2018mitigating}, for classification tasks, we pretrain a gender classifier and report the F1 score for gender prediction before and after debiasing to measure the degree of bias mitigation. For generation task, we also report the accuracy gap between a pretrained race classifier before and after debiasing. Additionally, for profession classification, \cite{ravfogel2020null} showed that the root-mean-square difference in the True Positive Rates between individuals (RMS TPR-GAP) with different gender is closely related to the Equal Opportunity fairness notion \cite{hardt2016equality}---hence we report this too.

\newpara{Rationale Faithfulness.} To ensure that extracted rationales are trustworhty, we evaluate faithfulness in rationale-based debiasing methods using comprehensiveness and sufficiency \citep{deyoung2020eraser}. Sufficiency measures the degree to which a rationale is adequate for making a prediction,
while comprehensiveness indicates whether all selections are necessary for making a prediction.
A smaller decrease in sufficiency and a larger decline in comprehensiveness indicate a high degree of faithfulness. 
We refer readers to 
\citep{deyoung2020eraser} for 
more details.
We also report the rationale selection ratio to measure conciseness of the extracted rationales.  


\section{Results and Analysis}
\subsection{Classification Tasks} 
\begin{table*}[t!]
\centering
\resizebox{0.9\textwidth}{!}{%
\begin{tabular}{cl}
\toprule
\begin{tabular}[c]{@{}c@{}} [-] Task \\ Rationale\end{tabular} &
  \begin{tabular}[c]{@{}l@{}}\green{Correct , Anderson .} Plowing through groups of innocent civilians is practiced by \green{islamic} \\ \green{terror groups} such as \green{ISIS} . It is also used by Palestinians to \green{kill babies} waiting at bus \\ stops in the arms of their \green{mother} .\end{tabular} \\\hdashline
\begin{tabular}[c]{@{}c@{}}Bias \\ Rationale\end{tabular} &
  \begin{tabular}[c]{@{}l@{}}\red{Correct , Anderson}. Plowing through groups of innocent civilians is practiced by islamic \\ terror \red{groups} such as ISIS . It is also used by Palestinians to kill \red{babies} waiting at bus \\ stops in the arms of their \red{mother} .\end{tabular} \\\hdashline
  \begin{tabular}[c]{@{}c@{}}[+] Task \\ Rationale (rerank)\end{tabular}  &
  \begin{tabular}[c]{@{}l@{}}Correct , Anderson . Plowing through groups of innocent civilians is practiced by \green{islamic} \\ \green{terror groups} such as \green{ISIS}. It is also used by Palestinians to \green{kill} babies waiting at bus \\ stops in the arms of their mother .\end{tabular} \\ \hdashline
\begin{tabular}[c]{@{}c@{}}[+] Task \\ Rationale (ours)\end{tabular}  &
  \begin{tabular}[c]{@{}l@{}}Correct , Anderson . Plowing through groups of innocent civilians is practiced by \green{islamic} \\ \green{terror groups} such as \green{ISIS}. It is also used by Palestinians to \green{kill} babies waiting at bus \\ stops in the arms of their mother .\end{tabular} \\ \midrule
  \midrule
  \begin{tabular}[c]{@{}c@{}}{[}-{]} Task\\ Rationale\end{tabular} &
  \begin{tabular}[c]{@{}l@{}}Showing solidarity with countries \green{inundated }with \green{refugees} by taking only \green{homosexuals} , \\ families and \green{orphans}. One \green{slip of} the \green{lip} and its \green{over} .\end{tabular} 
   \\\hdashline
\begin{tabular}[c]{@{}c@{}}Bias \\ Rationale\end{tabular} &
\begin{tabular}[c]{@{}l@{}}Showing solidarity with countries inundated with \red{refugees} by taking only \red{homosexuals} , \\ families and orphans . One slip of the \red{lip} and its over\end{tabular}
  \\\hdashline
\begin{tabular}[c]{@{}c@{}}{[}+{]} Task \\ Rationale (rerank)\end{tabular} &
  \begin{tabular}[c]{@{}l@{}}Showing solidarity with countries \green{inundated} with refugees by taking only homosexuals, \\ families and \green{orphans}. One \green{slip of} the lip and its \green{over} .\end{tabular} \\\hdashline
\begin{tabular}[c]{@{}c@{}}{[}+{]} Task \\ Rationale (ours)\end{tabular} &
  \begin{tabular}[c]{@{}l@{}}Showing solidarity with countries \green{inundated} with \yellow{refugees} by taking \green{only} \yellow{homosexuals} , \\ families and \green{orphans}. One slip of the lip and its over .\end{tabular} \\ 
\bottomrule
\end{tabular}%
}
\caption{Examples of extracted rationales in Toxicity Detection. Rationales used to predict toxicity are in green, those used to predict gender are in red, and overlap is in yellow. [-] indicates rationale generated before debiasing, and [+] indicates rationale generated after debiasing.  }
 \vspace{-0.8em}
\label{tab: example on toxicity}
\end{table*}

\paragraph{Dependence on sensitive information for task prediction.} First, we evaluate the appropriateness of the classification tasks by measuring how important tokens for task prediction are strong indicators of the sensitive information or bias. For toxicity detection, we observe in \cref{tab: entangle jigsaw} that when prediction models use only task rationales as input, they remain highly predictive for both the predictive task as well the bias prediction---showing minimal decrease in task and bias prediction performance when we switch from using full text input to only using task rationales as input (only 0.0005 points drop for toxicity detection, 0.0032 points drop for gender prediction). A similar phenomenon for profession classification, as seen in \cref{tab: entangle biobias}, 
indicates that both of these tasks might benefit from our debiasing method.



\paragraph{Performance of rationale-based debiasing methods.}
\cref{tab:result of classification variant } shows the comparison between our methods and other baseline along the dimensions of task performance, bias mitigation and rationale faithfulness. We achieve the maximum bias mitigation with the largest F1 score drop for gender (bias) prediction on both tasks (F1 drop of 0.1844 in toxicity detection and 0.6091 in profession classification). Secondly, debiasing affects minimally the task performance. We observed a minimal performance drop (0.00 for toxicity F1 and 0.01 for profession accuracy) after debiasing for our method whereas other methods with deabised rationales suffer from larger performance loss.
We see that debiasing constraint plays an important role during training to achieve better faithfulness, as we see our method achieves best comprehensiveness and sufficiency score.
Finally, our method achieves the best bias-performance trade-off by selecting sparser rationales as compared most of the other baselines. Rerank selects fewest tokens for rationales but such a sparse selection eventually hurts task performance. This also indicates a necessity of debiasing constraint at the training time rather than using it directly during inference. 


\paragraph{Performance of debiasing methods that do not produce rationales.}
We compare our algorithm with debaising algorithms that do not use rationales in \cref{tab:jigsaw debiasing baselines} and \cref{tab:biobias debiasing baselines} for both classification tasks.
We observe Adversarial Debiasing (Adv) achieves the maximum bias mitigation in both tasks. We argue that it debiases too much, to an extent that eventually hurts the task performance as we see large drops in toxicity F1 and profession accuracy. It is indicative that debiasing on the latent space leaves us with less room to control the balance between bias mitigation and task performance. Debiasing on embedding space (Embed) performs worse in the profession classification than other baselines that it not only harms task performance but also incorporates little debiasing. Upon investigation, we found that Embed uses word embeddings pre-trained on Google News. While the domain mismatch could lead the performance degradation for profession classification task (biographies being different than Google News); for toxicity detection the domain of online context matches with Embed pretraining and hence it attributes to the poor performance of the model itself. INLP is a strong baseline however it cannot produce any rationales hence lack transparency and control as compared to our method. 

\paragraph{Bias-performance trade-off.} We visualize the trade-off between  the degree of debiasing and task performance across various competing methods in \cref{fig: trade off in text classifition}. The upper-left corner indicates the optimal operational point. Among all other methods, we see that for both classification tasks, our method resides closest to the upper-left corner which confirms despite having stronger debiasing methods, we maintain the fair balance between task performance and the degree of debiaising. 


\subsection{Open-ended Generation Task} We present the comparative performances of the baselines and our method for the open-ended generation task in \cref{tab: Bold}.
While we see that debiasing in generation task is challenging as perplexity (PPL) for all methods are far from that of the ground-truth human-written answers, 
our method achieves the best bias mitigation as well as best perplexity and BertScore as compared to other debiasing methods. While PPLM is fluent with a good perplexity and mitigates bias reasonably, it has low BertScore indicating low generation quality. 
We achieve better generation results by using sparser rationales as compared to GPT2 and Probability baselines. While Rerank selects fewest input words as rationales it eventually have poor generation quality showing lack of control on bias exposure to maintain task performance. While the Probability model acted as a strong baseline for classification tasks, for generation task, it performs worse than the GPT2 baseline. We attribute this to the lack of independence assumption between $p(z^b_i=0)$ and $p(z^t_i=0)$, as task labels and bias labels appears to be closely related and hence directly minimizing their sum in $D$ might suffer from confounding in some cases. We also notice that both PPLM and our method achieve best faithfulness in terms of sufficiency but we achieve that using sparser rationales and better generation quality.


\subsection{Case Study}
We compare extracted rationales with two different inputs across different rationale-based debiasing methods for toxicity detection task in \cref{tab: example on toxicity}. More examples are provided in the Appendix~\ref{sec: more example}. 


In the first example, `mother' appears to be in the task rationales for toxicity as often offensive expressions and slangs include the word `mother'. On the other hand, `mother' is also highly predictive of gender (female). However, in the current context, `mother' is not indicative of toxicity but only acts as a sensitive token, hence our method penalizes its importance and does not use it for the task prediction after debiasing.

In the second example, `lip' (frequently appears as a part of \emph{lipstick}) and `homosexuals' appear as indicator for gender as well as predicting toxicity. It is understandable that `homosexuals' strongly indicates toxicity as it regularly appears in homophobic  comments. While removing both them will decrease gender bias greatly, something that happens for Rerank baseline, it is not \emph{fair} to not include `homosexuals' in task rationales. While our method drops `lip' from task rationales after debiasing it still keeps (and fairly so) `homosexuals' in its task rationales thus controlling the bias exposure for a fair and interpretable toxicity prediction.

\section{Conclusion}
We proposed a fair and interpretable debiasing method that can control bias exposure by balancing bias mitigation and task performance. While previous methods often debias too strongly or with lesser control and transparency, we show, on three different tasks, that our method achieves the best trade-off between task performance and bias mitigation, while producing the most faithful rationales for the debiased task prediction. We also indicate cases where it is even necessary to keep sensitive information that is useful for task output. Our model provides fair control on bias exposure, especially in such cases, instead of blindly debiasing the input with minimal interpretation.

\section{Limitations}
It is often a delicate decision that how much a biased token contributes to the original predictive task. Especially on tasks such toxicity detection, sentiment analysis, it is common to see the mentions of minority groups (Example 2 in \cref{tab: example on toxicity}) that carry pivotal information for the original task label (in our example, `toxic'). Hence, it is inevitable, at the surface, to include those mentions in order to maintain task performance. Therefore, we allow models to use biased words when necessary, but only in conjunction with immediate notifications sent to users, asking for reconsideration or revision of the input before using them in public. When possible, we adjust the contribution of biased tokens to their existing unbiased replacements. However, we unable to `generate' an unbiased replacement when a suitable one is not present in the current input. As a result, complete debiasing can be achieved by involving humans in the loop so that a \emph{better} alternative is found and used.

Another possible concern would be the usage of sensitive information. It is worth mentioning that in this work, we focus on controlling bias exposure to maintain a balance between debiasing and task performance with an explanation instead of removing all sensitive information as a process of debiasing. However, as a special case of our system, it is possible to set the bias threshold to a minimal value which results in removing all biased tokens, prohibiting using any sensitive information. Although, this may affect the task performance considerably which is a trade-off the end-user has to consider.
 
\section{Ethical Considerations}
Efforts have been made in the last few years to develop artificial intelligence systems that are fairness-aware to prevent different types of bias. Nevertheless, a malicious user could potentially abuse the system in an adversarial manner. It is possible to preserve highly-biased parts of the input by optimizing our debiasing constraint in a reversed way, which could be used as harmful input for downstream tasks, causing undesired ethical implications. It is necessary and desirable to conduct sanity auditing by all the stakeholders. Our recommendation is that users who deploy our system should also provide a visualization of the generated `debiased' rationale (similar to \cref{tab: example on toxicity}), in order to facilitate the verification process.   

\section*{Acknowledgements} We thank Taylor Berg-Kirkpatrick,
Yuheng Zhi, and anonymous reviewers for providing valuable feedback. BPM is partly supported by an Adobe Research Fellowship, a Friends of the International Center Fellowship--UC San Diego, NSF Award \#1750063, and MeetElise. 

\bibliography{anthology}
\bibliographystyle{acl_natbib}

\clearpage
\appendix


\section{Implementation Details}
\label{sec: appendix}
\textbf{Classification Tasks}. In order to segment words in the sentences, we utilize the popular \href{https://www.nltk.org/_modules/nltk/tokenize.html#word_tokenize}{ \texttt{nltk.tokenize.word\_tokenizer}} from \texttt{nltk} package, and choose GLoVe \citep{glove} as our word embeddings. We choose to use bidirectional LSTM as the extractor, with hidden dimension as 150. Then we build another bidirectional LSTM with the same dimension on top of extractor as the classifier. We first pretrain a bias extractor and classifier with the above structures. During the training process, we set the selection ratio as 0.5 (this number does not matter according to our experiments. The intuition is that Kuma will change the prediction globally according to the selection ration. Then we only need to adjust the threshold $A$ in constraint $D$ 
to obtain compatible results.) Then with the energy given by this bias REF, we can calculate the debiasing constraint to update the task REF. In implementation process, we set  LASSO weight to be 0 and set the selection rate as 0.7 for both toxicity detection and profession classification. We also tried with other weights (0.01, 0.1, etc) and no significant change is observed. 

\paragraph{Generation Tasks.} The backbone of this task is GPT2 (117M parameters) open-sourced in huggingface\footnote{https://huggingface.co/gpt2}. The bias extractor, bias classifier and task extractor are the same as in the classification tasks except that we use GPT2 tokenizer and word embeddings for bias extractor. However, instead of using task classifier, we put GPT2 on top of the task extractor. The tokenizer and word embeddings for the task extractor are also from GPT2. If some words are not selected, then we multiply zero on the corresponding word embeddings before GPT-2 process them. 
For the whole training procedure, We first pretrain GPT2 on the whole BOLD dataset; then we also pretrain the bias REF with the prompts as the input and the bias labels as the output. After that, we train our task rationale extractor with GPT2 fixed. We guarantee there is no data overlap between any training/validation/test set.

\paragraph{Details about Metrics}
For classification task, our F1 and accuracy scores are calculated with standard \href{https://scikit-learn.org/stable/modules/classes.html#module-sklearn.metrics}{ \texttt{sklearn.metrics}} from \texttt{sklearn} package. For generation task, we calculate PPL and BertScore with official \href{https://huggingface.co/docs/evaluate/index}{\texttt{evaluate}} pacakge from Huggingface. 

\paragraph{Resources} 
The whole experiments are run on eight 3090Ti GPUs with 24G DRAM. 
All the examples are run on single GPU. It takes about eight hours for the model trained in toxicity detection task and profession classification task to converge. Then as for the open-ended generation task,  finetuning a pretrained GPT2 from Huggingface takes around two hours and the pretraining of bias REF takes about one hour. The training of task REF takes around another one hour, which means the whole process for one setting takes about 4 hours. 

\section{Hyperparameter Study}
\label{sec: Hyperparameter Study}
In this section, we explore the effects of the hyperpameter: threshold $A$ in constraint $D$
and the selection ratio. The results are reported in Table \ref{tab:hyperparameter_study}. From the table, we could observe (1) The debiasing results are usually better when bias threshold $A$ is around $-\text{log}(1-0.5)$. This observation is not surprising. Imagine the extreme cases, if $A = -\text{log}(1 - 1.0) = +\infty$, then $D(i)$ will consistently be $0$, contributing nothing to the objective, Then if $A = -\text{log}(1 - 0.0) = 0$. Then the outcome energy on every word will be penalized, including both biased words and unbiased words, leading to degenerated performances.  (2) The performances of the prediction on Toxicity is not very sensitive to the parameter $A$, but the selection ratio has much larger influences. It is also intuitive since we can always make better predictions with more input of the text, \emph{i.e.}, larger selection ratio.

\begin{table}[ht]
    \centering
    \resizebox{\linewidth}{!}{%
    \begin{tabular}{cccc}
    \toprule
    Selected & Threshold $A$ & Toxicity F1$\uparrow$ & Gender F1$\downarrow$\\
    \midrule
    0.7 & $-\text{log}(1 - 0.3)$ & 0.6417 & 0.2837 \\
    0.7 & $-\text{log}(1 - 0.5)$ & 0.6522 & 0.3115 \\
    0.7 & $-\text{log}(1 - 0.7)$ & 0.7255 & 0.3723 \\
    0.5 & $-\text{log}(1 - 0.3)$ & 0.6459 & 0.2103 \\
    0.5 & $-\text{log}(1 - 0.5)$ & 0.6192 & 0.2026 \\
    0.5 & $-\text{log}(1 - 0.7)$ & 0.6205 & 0.2219 \\
    0.3 & $-\text{log}(1 - 0.3)$ & 0.4634 & 0.1826 \\
    0.3 & $-\text{log}(1 - 0.5)$ & 0.4633 & 0.1751\\
    0.3 & $-\text{log}(1 - 0.7)$ & 0.4632 & 0.1803 \\
     \bottomrule
    \end{tabular}}
    \caption{Hyperparameter Study}
    \label{tab:hyperparameter_study}
\end{table}

\begin{table*}[htbp]
\centering
\resizebox{0.9\textwidth}{!}{%
\begin{tabular}{cl}
\toprule
\begin{tabular}[c]{@{}c@{}}[-] Task \\ Rationale\end{tabular} &
  \begin{tabular}[c]{@{}l@{}}\green{Trump's insults everyone}; he believes we are so \green{ignorant} we'll believe anything he says and \\ miss the \green{contradictions}. He doesn't even bother with coherent speeches, he just \green{mouths} \\ some words and listens to the cheers. There must be a \green{disconnect} between the ears and \\ brains \green{ of the women} who hear The \green{Donald's put-downs and swoon}. Or maybe they just \\ think it applies to all other \green{'fat , ugly bimbos '} and not themselves .\end{tabular} \\\hdashline
\begin{tabular}[c]{@{}c@{}}Bias \\ Rationale\end{tabular} &
  \begin{tabular}[c]{@{}l@{}}\red{Trump's} insults everyone; \red{he} believes we are so \red{ignorant} we'll believe anything he says and \\ miss the \red{contradictions}. \red{He} doesn't even bother with coherent speeches, \red{he} just mouths \\ some words and listens to the cheers. There must be a disconnect between the ears and \\ brains of the \red{women} who hear The \red{Donald's} put-downs and swoon. Or maybe they just \\ think it applies to all other \red{'fat , ugly bimbos '} and not themselves .\end{tabular} \\ \hdashline
\begin{tabular}[c]{@{}c@{}}[+] Task \\ Rationale\\(rerank) \end{tabular} &
  \begin{tabular}[c]{@{}l@{}}Trump's \green{insults everyone}; he believes we are so ignorant we'll believe anything he says and \\ miss the contradictions. He doesn't even bother with coherent speeches, he just \green{mouths} \\ some words and listens to the cheers. There must be a \green{disconnect} between the ears and \\ brains of the women who hear The Donald's \green{put-downs and swoon}. Or maybe they just \\ think it applies to all other 'fat , ugly bimbos ' and not themselves .\end{tabular} \\ \hdashline
\begin{tabular}[c]{@{}c@{}}[+] Task \\ Rationale \\(ours) \end{tabular} &
  \begin{tabular}[c]{@{}l@{}}Trump's \green{insults everyone}; he believes we are so \yellow{ignorant} we'll believe anything he says and \\ miss the contradictions. He doesn't even bother with coherent speeches, he just \green{mouths} \\ some words and listens to the cheers. There must be a \green{disconnect} between the ears and \\ brains of the women who hear The Donald's \green{put-downs and swoon}. Or maybe they just \\ think it applies to all other 'fat , \yellow{ugly bimbos} ' and not themselves .\end{tabular} \\ 
\bottomrule
\end{tabular}%
}
\caption{\small Debiasing Example in Toxicity Detection. Task rationales are in green, bias rationales are in red, and overlap is in yellow. [-] indicates rationale generated without debiasing, and [+] indicate that with debiasing.}
\label{tab: additional example on toxicity}
\end{table*}

\section{Criteria of Selecting Reference Model}
Here we provide results on our reference model in toxicity detection (shown in \cref{tab: jigsaw why reference (gender)}) and in profession classification (shown in \cref{tab: biobias why reference (gender)}). From the tables, we found the predicting gender on gender rationales have almost same performance with that on full text, which confirms that the reference model in each experiment are good enough to generate high-quality rationale used in debiasing constraint. 


\begin{table}[htbp]
\begin{tabular}{@{}ccc@{}}
\toprule
    & Gender Accuracy & Gender F1
    \\ \midrule
full text        & 0.87 & 0.55                      \\
gender rationale & 0.87 & 0.53                    \\ \bottomrule
\end{tabular}
\caption{The gender predict performance of the pretrained reference model. The required selection rate is no more than 50\% (Jigsaw)}
\label{tab: jigsaw why reference (gender)}
\end{table}


\begin{table}[htbp]
\begin{tabular}{@{}ccc@{}}
\toprule
    & Gender Accuracy & Gender F1
    \\ \midrule
full text        & 0.98 & 0.99                      \\
gender rationale & 0.98 & 0.99                    \\ \bottomrule
\end{tabular}
\caption{The gender predict performance of the pretrained reference model. The required selection rate is no more than 50\% (BioBias)}
\label{tab: biobias why reference (gender)}
\end{table}

\section{Additional Debiasing Example}
\label{sec: more example}
We provide another debiasing example from the task Toxicity Detection in Table \ref{tab: additional example on toxicity}. From the example, we found that the commentor is criticizing Donald Trump. 
Trump is marked as toxic token,due to the strong correlation of sentence mentioning Trump and a toxic label in the dataset. However, they are also gendered words, as Donald Trump is a well-known male. Debiasing can help to delete the biased words that are not absolutely necessary for making a task prediction. However, for words like `ignorant' and `ugly bimbos', though they are highly predictable for gender (due to the frequent co-appearance), they are necessary parts for a sentence being toxic. 

\end{document}